%% file: paper.tex
\documentclass[]{bytedance_seed}

\input{macros}

\title{GR-Dexter Technical Report}

\author[]{ByteDance Seed}

\contribution[]{Full Author List in \hyperref[sec:contributions]{Contributions and Acknowledgements}}

\abstract{

Vision-language-action (VLA) models have enabled language-conditioned, long-horizon robot manipulation, but most existing systems are limited to grippers. Scaling VLA policies to bimanual robots with high degree-of-freedom (DoF) dexterous hands remains challenging due to the expanded action space, frequent hand-object occlusions, and the cost of collecting real-robot data.
We present \name{}, a holistic hardware-model-data framework for VLA-based generalist manipulation on a bimanual dexterous-hand robot. Our approach combines the design of a compact 21-DoF robotic hand, an intuitive bimanual teleoperation system for real-robot data collection, and a training recipe that leverages teleoperated robot trajectories together with large-scale vision-language and carefully curated cross-embodiment datasets.
Across real-world evaluations spanning long-horizon everyday manipulation and generalizable pick-and-place, \name{} achieves strong in-domain performance and improved robustness to unseen objects and unseen instructions. We hope \name{} serves as a practical step toward generalist dexterous-hand robotic manipulation.

}

\date{\today}

\checkdata[Correspondence]{wenruoshi@bytedance.com}

\checkdata[Project Page]{\url{https://byte-dexter.github.io/gr-dexter}}

\begin{document}
\maketitle

\input{sections/introduction}

\input{sections/hands_system}

\input{sections/gr_dexter_model}

\input{sections/experiments}

\input{sections/related_works}

\input{sections/limitation_future_work}

\clearpage
\bibliographystyle{plainnat}
\bibliography{main}
\input{sections/contribution}

\clearpage
\beginappendix
\input{sections/appendix}

\end{document}

%% file: macros.tex
\usepackage{minitoc}
\usepackage{url}
\usepackage{amssymb}
\usepackage[utf8]{inputenc}
\usepackage{microtype}
\usepackage{booktabs}
\usepackage{pifont} 
\usepackage{multirow}
\usepackage{makecell}
\usepackage{paralist}
\usepackage{xspace}
\usepackage{color}
\usepackage{xcolor}
\usepackage{colortbl}
\usepackage{adjustbox}
\usepackage{hyperref} 
\usepackage[edges]{forest}
\usepackage{tikz} 
\usepackage{caption}
\usepackage{amsfonts}
\usepackage[toc,page,header]{appendix}
\usepackage[utf8]{inputenc}
\usepackage{bbm}
\usepackage{enumitem}
\usepackage{wrapfig}

\definecolor{seedc}{RGB}{7, 92, 173}

\newcommand{\name}[1]{GR-Dexter}

\newcommand{\hardware}[1]{ByteDexter}

\newcommand{\ba}{\mathbf{a}}

\renewcommand{\paragraph}[1]{\vspace{0.1em}\noindent\textbf{#1}}

%% file: sections/introduction.tex
\section{Introduction}
\label{sect:intro}
Generalist manipulation policies powered by vision-language-action (VLA) models have enabled language-conditioned control and long-horizon instruction following in robot manipulation~\cite{kim2024openvla, wen2025dexvla, gr00tn1_2025, gr3_2025}. However, most existing policies are deployed on bimanual robots with gripper-based end effectors. Extending these capabilities to robots equipped with dexterous hands remains underexplored. As robots move toward general-purpose operation in cluttered, human-centered environments, dexterous hands hold greater potential for achieving human-level manipulation. Yet this promise comes with substantial challenges: high degree-of-freedom (DoF) hands expand the control space by dozens of DoFs, while introducing perception difficulties---frequent occlusions between fingers and between the hand and target objects. Moreover, as a data-driven paradigm, VLA performance depends critically on the quality and diversity of robot trajectories for dexterous bimanual manipulation.

\begin{figure}[h]
    \centering
        \centering
        \includegraphics[width=0.9\linewidth]{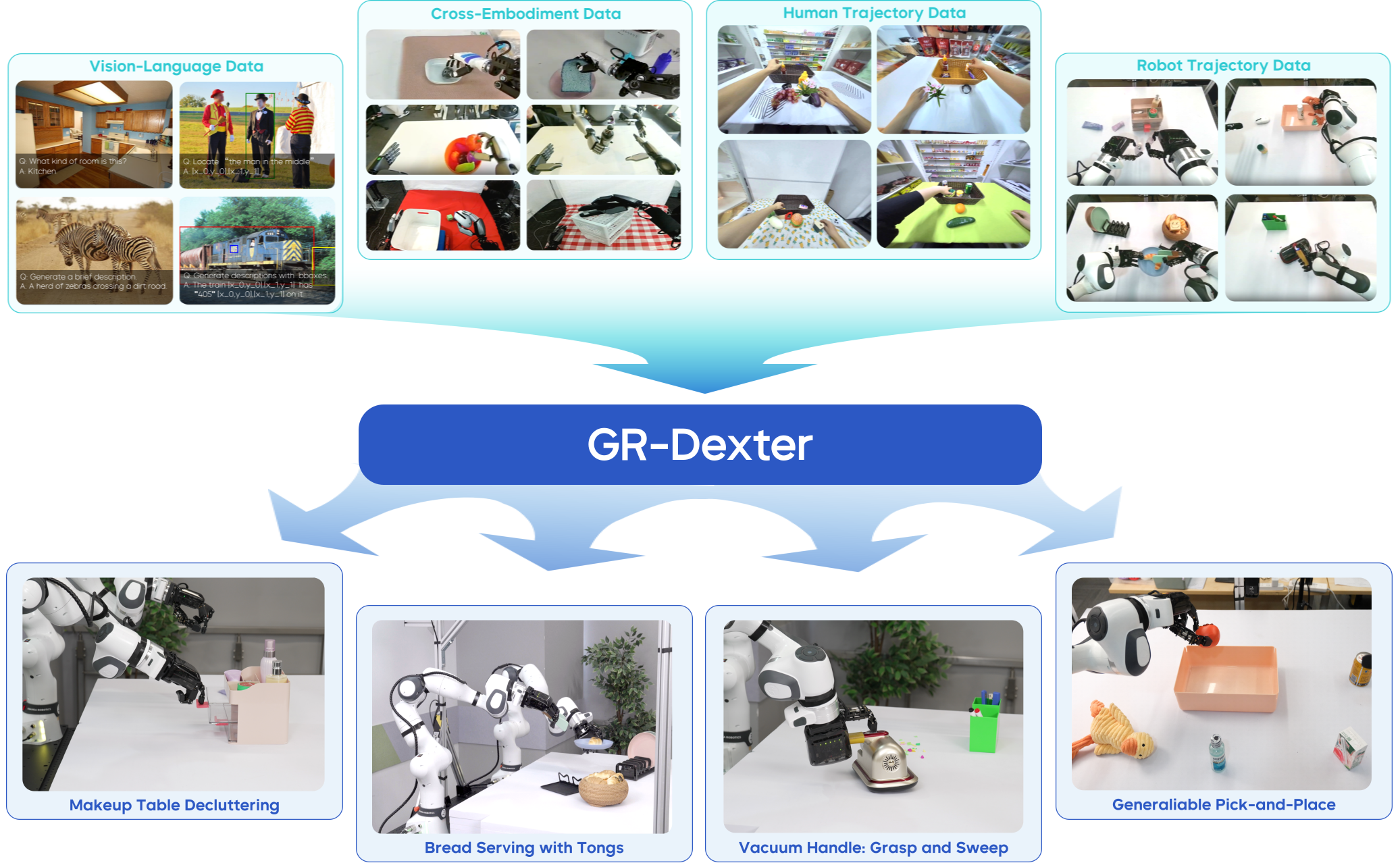}
        \caption{
        GR-Dexter performs dexterous long-horizon daily tasks and generalizes to out-of-domain settings by learning from four data sources: vision--language, cross-embodiment, human-trajectory, and robot-trajectory data.
        }

        \label{fig:teaser}
\end{figure}

We present the ByteDexter V2 hand---a 21-DoF linkage-driven anthropomorphic robotic hand, designed as a self-contained, modular end-effector for dexterous manipulation, with space,  complexity, and maintainability as key design constraints. Compared with ByteDexter V1~\cite{wen2025dexterous} and ILDA hand~\cite{KimILDA2021}, V2 adds an additional thumb DoF while further reducing overall size. With actuators integrated within the palm, the hand achieves a compact form factor (219~mm height, 108~mm width). It also incorporates high-density piezoresistive tactile sensors at the fingertips. 

We then introduce a VLA model \name{} and training recipe tailored to a \textbf{56-DoF} bimanual system equipped with ByteDexter V2 hands. The policy is built on a pre-trained VLM~\cite{bai2025qwen25vltechnicalreport}, and is co-trained on a mixture of data sources, including teleoperated robot trajectories, vision-language data, cross-embodiment demonstrations, and human trajectories. Because bimanual arm teleoperation is challenging even with simple grippers, efficient demonstration collection becomes more difficult when each end-effector is a 21-DoF dexterous hand. We address this challenge with a teleoperation interface comprised of a Meta Quest headset and Manus gloves, which retarget tracked human wrist poses and hand motions to joint position commands in real time.

Beyond collecting teleoperated robot trajectories, anthropomorphic high-DoF hands also provide a promising data-scaling path: the structural similarity between human and robot hands makes it feasible to directly leverage large-scale egocentric hand-object interaction datasets that cover diverse everyday dexterous behaviors~\cite{grauman2022ego4d, banerjee2024hot3d, hoque2025egodex, grauman2024ego}. Our teleoperation pipeline then enables efficient collection of a small amount of on-robot data for fine-tuning the pretrained models to adapt to the target platform.

\begin{figure}[!htbp]
    \centering
    \begin{subfigure}{0.87\textwidth}
        \centering
        \includegraphics[
            width=\linewidth,
            trim=6mm 5mm 6mm 5mm, %
            clip
        ]{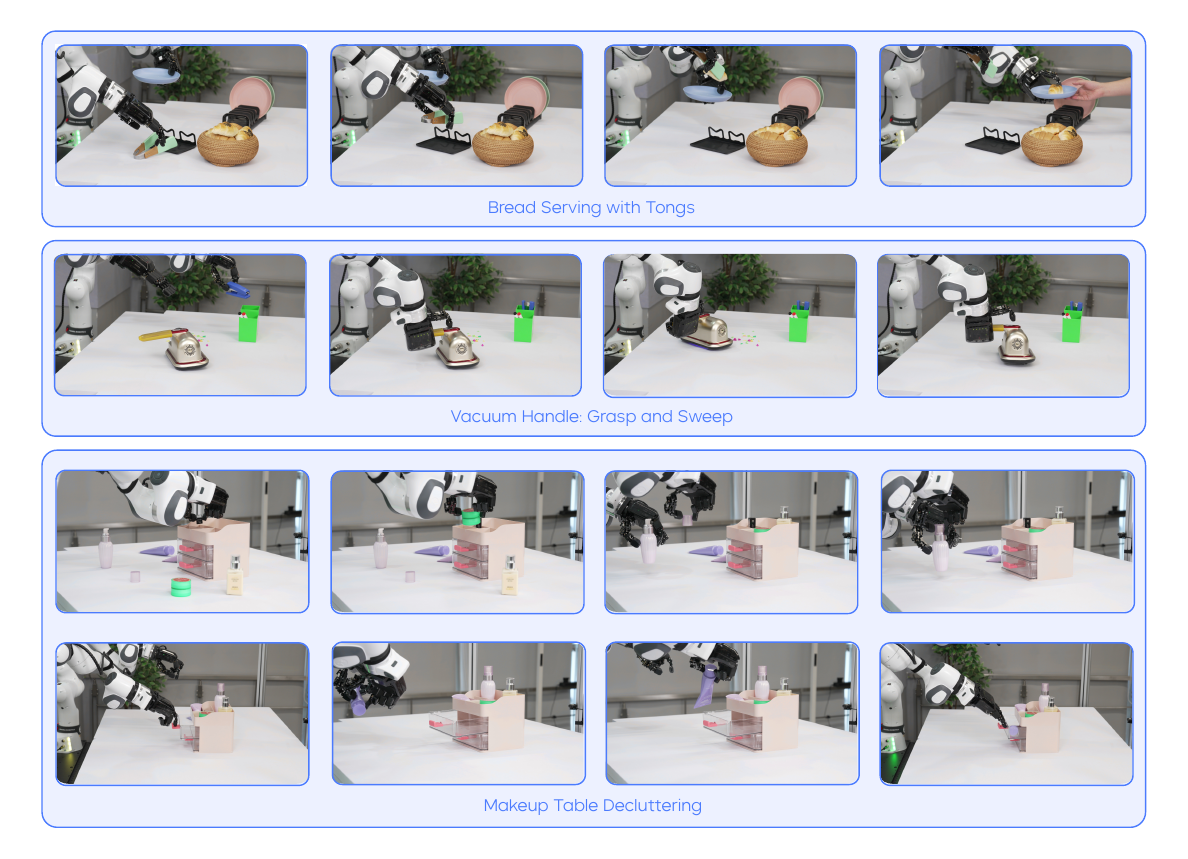}
        \caption{\name{} performs long-horizon tasks.}
        \label{fig:long_horizon}
    \end{subfigure}
    
    \begin{subfigure}{0.87\textwidth} 
        \centering
        \includegraphics[
            width=\linewidth,
            trim=6mm 5.5mm 6mm 3mm, %
            clip
        ]{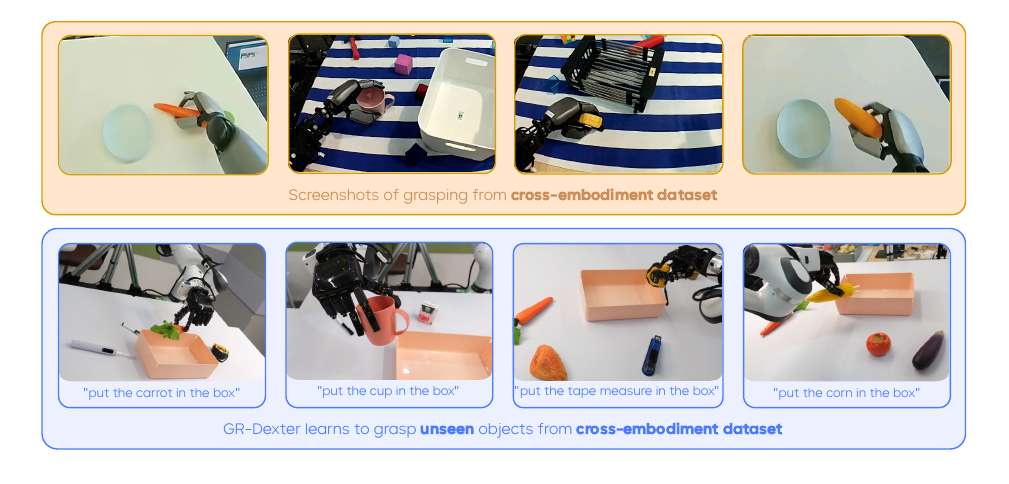} 
        \caption{\name{} is capable of grasping unseen objects.}
        \label{fig:cross_embodied}
    \end{subfigure}

    \begin{subfigure}{0.87\textwidth} 
        \centering
        \includegraphics[
            width=\linewidth,
            trim=6mm 1mm 6mm 1mm, %
            clip
        ]{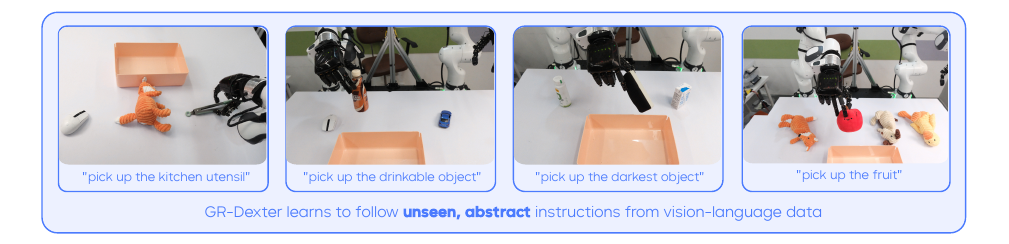} 
        \caption{\name{} follows unseen language instructions.}
        \label{fig:unseen_ins}
    \end{subfigure}

    \caption{\textbf{Capabilities.} \name{} robustly completes long-horizon daily tasks. It also learns to grasp unseen objects, and follow unseen, abstract language instructions.}
    \label{fig:capabilities}
\end{figure}

We evaluate \name{} in real-world experiments across two task categories: (1) long-horizon manipulation, and (2) generalizable pick-and-place. Results show strong performance in both in-domain settings and challenging unseen scenarios, including novel objects and previously unseen language instructions (Fig.~\ref{fig:capabilities}). This performance stems from co-training with large-scale vision-language data, cross-embodiment data, and human trajectories, which preserves robust grasping behaviors on in-domain sub-tasks while improving generalization to out-of-distribution (OOD) cases. Moreover, \name{} successfully completes long-horizon everyday tasks, highlighting its practical bimanual dexterity in the real world.

%% file: sections/hands_system.tex
\newpage

\begin{figure}[h]
    \centering
    \begin{subfigure}{0.5\textwidth} 
        \centering
        \includegraphics[width=\linewidth]{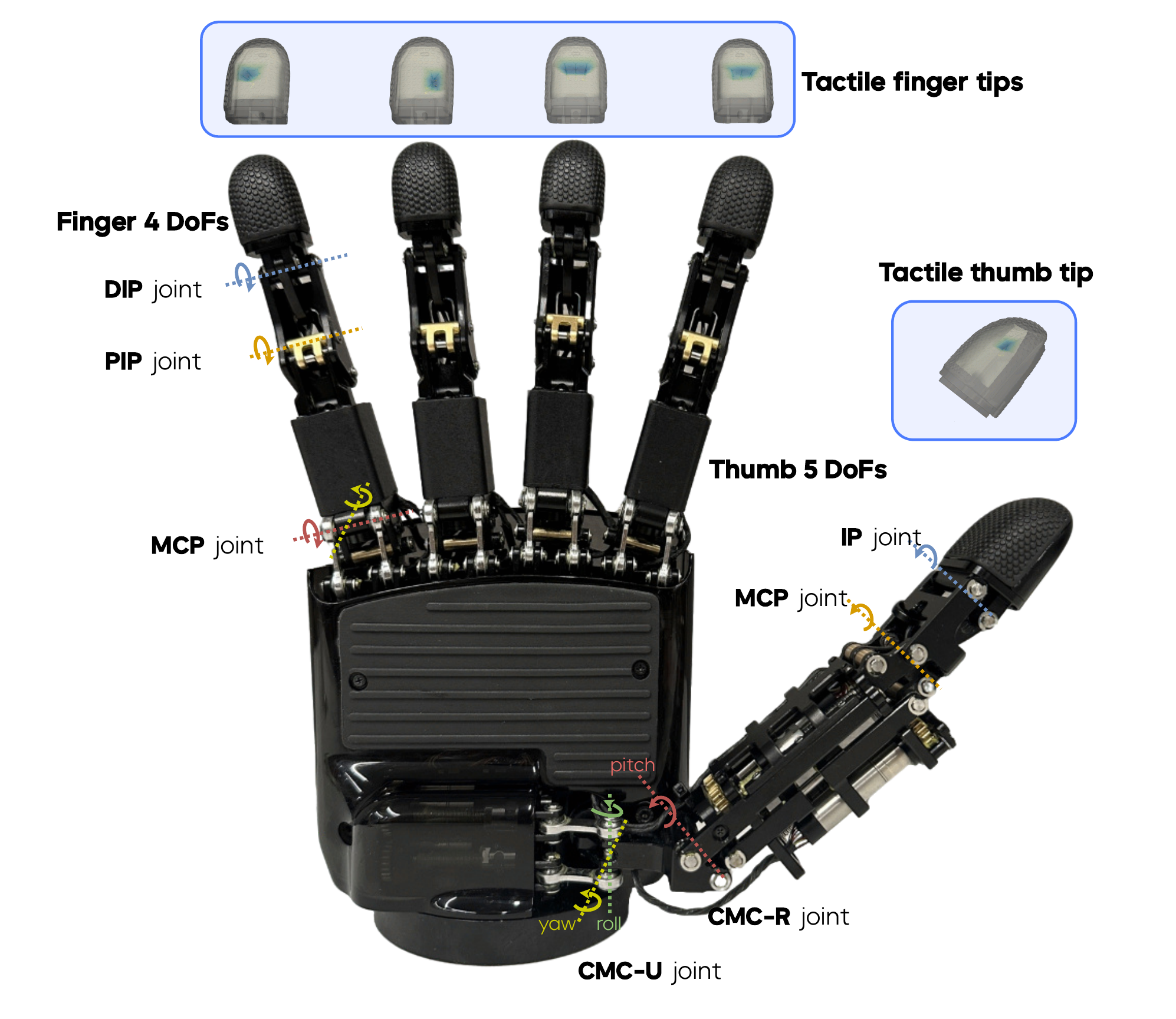} 
        \caption{ByteDexter V2 DoF distribution and tactile sensors.}
        \label{fig:hand}
    \end{subfigure}
    \begin{subfigure}{0.45\textwidth} 
        \centering
        \includegraphics[width=\linewidth]{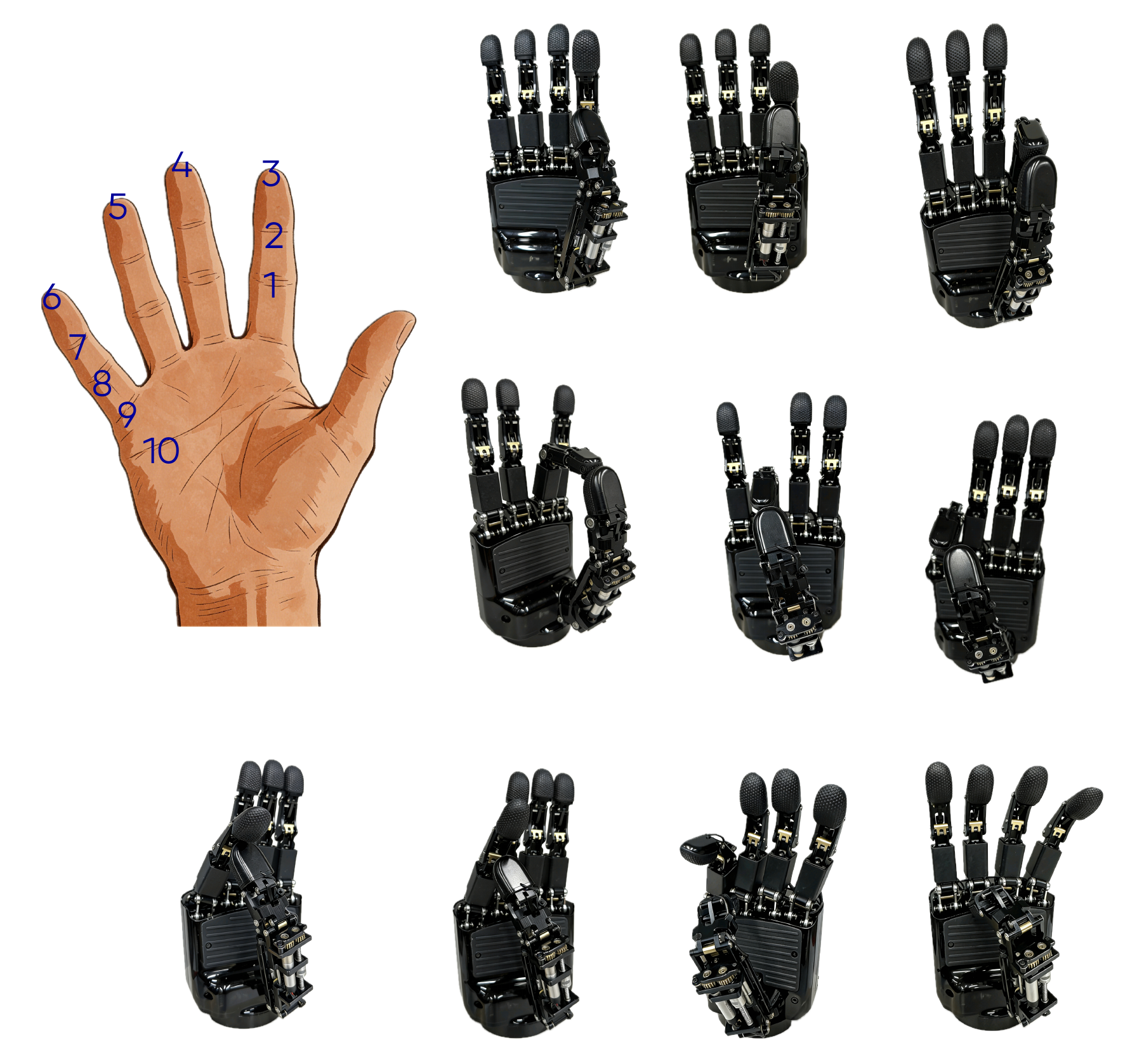} 
        \caption{ByteDexter V2 scores 10 in the Kapandji test}
        \label{fig:hand_kapandji}
    \end{subfigure}
    \caption{\textbf{The ByteDexter V2 hand.} We show the DoF distribution, tactile fingertips, and the thumb's opposition capability.}
    \label{fig:hand-design}
\end{figure}
\section{ByteDexter Robotic Hand}
\label{sec:hands_system}

The ByteDexter hand series employ a linkage-driven transmission mechanism for its advantages in force transparency, durability, and ease of maintenance. As an upgraded successor to the V1 hand \cite{wen2025dexterous}, the \textbf{ByteDexter V2 hand} introduces an additional thumb DoF, bringing the total to 21 DoFs, while simultaneously reducing the overall hand size (height: 219mm, width: 108mm). Each finger has four DoFs, and the thumb has five, providing a wider range of oppositional motions, illustrated in Fig.~\ref{sec:hands_system}. We also demonstrate its human-like grasping capability by executing all 33 Feix grasp types \cite{feix2016} (Appendix Fig.~\ref{fig:hand_feix}). 

\subsection{Hand Design}

\textbf{Fingers (index, middle, ring, little)} The four fingers share a modular architecture. Each finger comprises a universal joint at the MCP (metacarpophalangeal) and two revolute joints at the PIP (proximal interphalangeal) and DIP (distal interphalangeal). The two DoFs at the MCP are actuated by two motors housed in the palm, enabling abduction--adduction and flexion--extension. Unlike the ILDA hand~\cite{KimILDA2021}, ByteDexter V2 decouples PIP flexion from MCP flexion, so that the PIP is independently actuated by a dedicated third motor.

\textbf{Thumb} In the human hand, the saddle-shaped carpometacarpal (CMC) joint enables flexion--extension and abduction--adduction, which are critical for dexterous in-hand manipulation. ByteDexter V2 employs a universal joint at the CMC together with an additional revolute joint to approximate these kinematics and preserve key functional characteristics (Fig.~\ref{fig:hand}). The compact, integrated thumb mechanism minimizes internal volume while substantially increasing the thumb’s range of motion. The resulting enlarged reachable workspace enables robust oppositional contact with all four fingers (Fig.~\ref{fig:hand_kapandji}).

\begin{figure}[h]
    \centering
        \centering
        \includegraphics[width=0.8\linewidth]{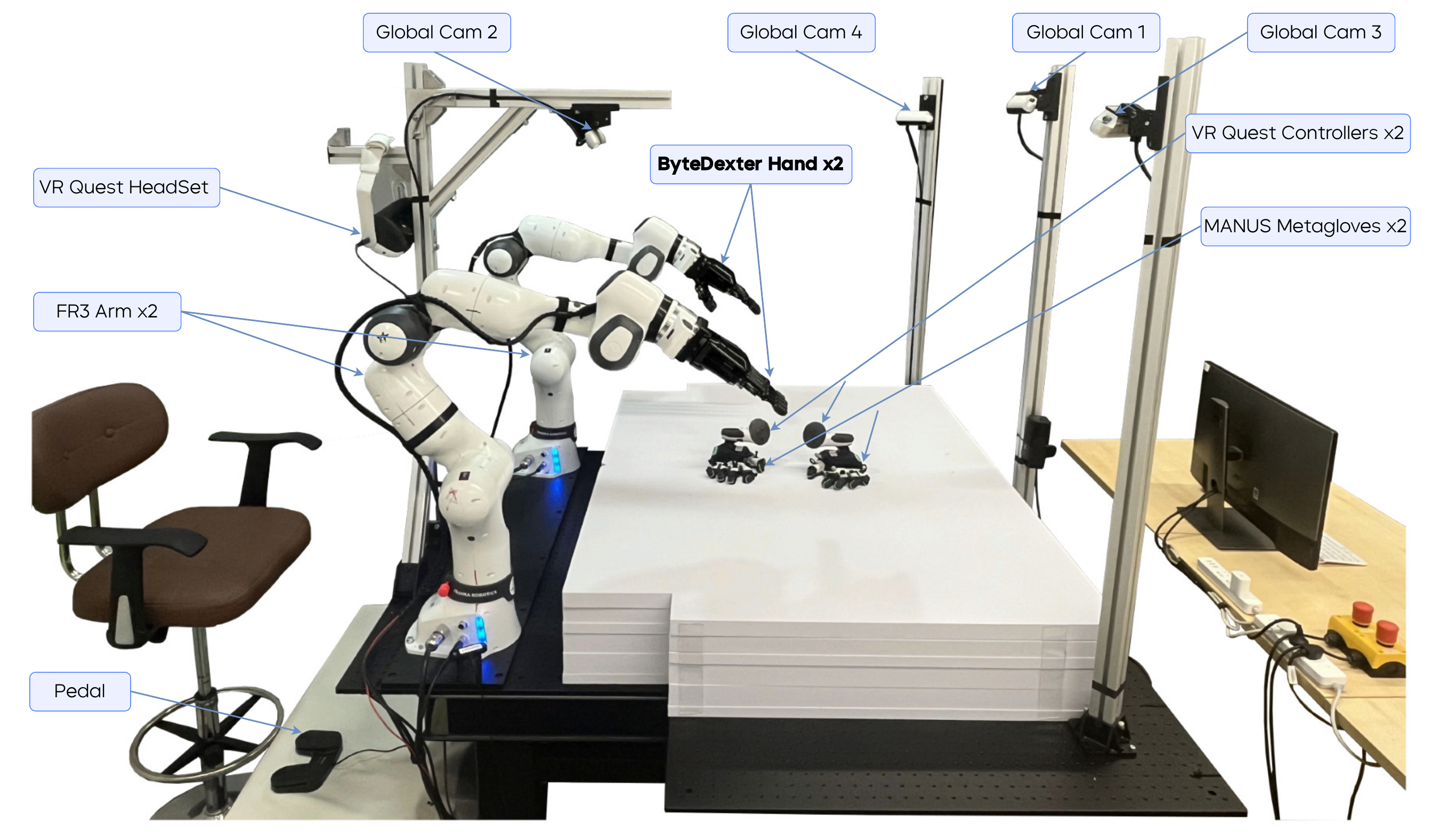}
        \caption{The bimanual robotic system comprising two Franka Research 3 arms equipped with ByteDexter V2 hands. Data are collected via a teleoperation interface using a Meta Quest VR headset, Manus gloves with mounted VR tracking controllers, and a set of global RGB-D cameras.}
        \label{fig:teleop_setup}
\end{figure}

\textbf{Underactuation} The DIP joints of four fingers and the IP (interphalangeal) joint of the thumb are underactuated. ByteDexter V2 implements a biomimetic four-bar linkage mechanism that couples each DIP to its corresponding PIP, reproducing the intrinsic kinematic coupling observed in the human DIP–PIP joint complex.

\textbf{Tactile Sensing} The five fingertips of ByteDexter V2 are covered with high-density piezoresistive tactile arrays that measure normal contact forces (Fig.~\ref{fig:hand}). The visualization encodes contact location and force magnitude, and the arrays provide fine spatial resolution over the fingertip, finger pad, and lateral surface.

\subsection{Bimanual System and Control}

We built a dual-arm platform equipped with two ByteDexter V2 hands for bimanual manipulation (Fig.~\ref{fig:teleop_setup}). The resulting 56-DoF robot is designed to support coordinated arm-hand control for reliable dexterous grasping and manipulation. To mitigate occlusions and capture hand-object interactions from multiple views, we deploy four global RGB‑D cameras: one primary egocentric view and three complementary third-person views. The platform supports both teleoperated data collection and autonomous policy rollouts.

\paragraph{Bimanual teleoperation}
We collect real-world robot data using a bimanual teleoperation interface consisting of a Meta Quest VR setup for tracking wrist poses, two Manus Metagloves for capturing hand movements, and foot pedals to enable/disable teleoperation. Two Meta Quest controllers are mounted on the dorsal side of the gloves to improve the reliability of coordinated wrist–hand tracking. This setup allows teleoperators to simultaneously coordinate two Franka arms together with two ByteDexter V2 hands during long-horizon manipulation tasks. Human motions are retargeted in real time to joint position commands via a whole-body controller, providing a kinematically consistent mapping. The system incorporates safety mechanisms to handle intermittent visual tracking loss and mitigate hazardous operation. Hand-motion retargeting is formulated as a constrained optimization problem that combines wrist-to-fingertip and thumb-to-fingertip alignment terms with collision-avoidance constraints and regularization, and is solved using Sequential Quadratic Programming.

\paragraph{Policy rollout} During policy rollout, our model generates future action chunks that promote coordinated, temporally consistent arm–hand motions for dexterous manipulation. The parameterized trajectory optimizer smooths the generated actions, which is critical for delicate grasping, and ensures smooth transitions both within and across chunks.

\begin{figure}[h]
    \centering
        \centering
        \includegraphics[width=1.0\linewidth]{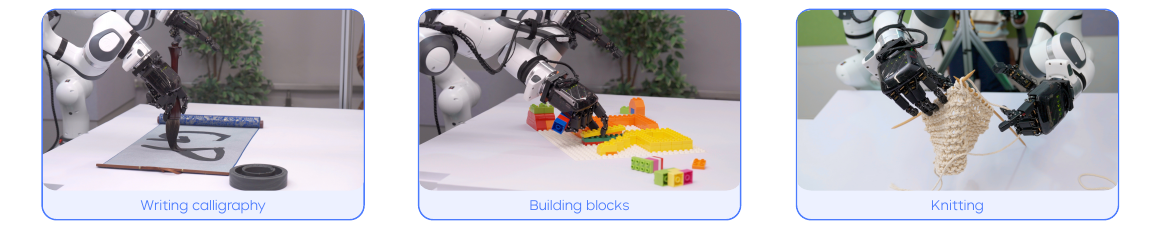}
        \caption{Teleoperation capability in long-horizon dexterous grasping and bimanual manipulation tasks.}
        \label{fig:teleop_demo}
\end{figure}

The bimanual system demonstrates efficiency, human-like dexterity, and reliable long-duration operation. After minimal training, teleoperators successfully completed tasks ranging from coarse manipulation (e.g., building blocks) to fine motor tasks (e.g., knitting), as shown in Fig.~\ref{fig:teleop_demo}. The breadth of tasks highlights the system’s suitability for real-world bimanual manipulation, enabling reliable data collection and policy evaluation.

%% file: sections/gr_dexter_model.tex
\section{The GR-Dexter Model}
\label{sec:GR-Dexter Model}

\name{} follows GR-3~\cite{gr3_2025} and adopts a Mixture-of-Transformer architecture for a vision-language-action (VLA) model $\pi_{\theta}$ of 4B parameters. $\pi_\theta(\ba_t\mid l, \mathbf{o}_{t}, \mathbf{s}_{t})$ controls a bi-manual robot with fixed base by generating a $k$-length action chunk $\ba_t  = a_{t:t+k}$ conditioned on the input language instruction $l$, observation $\mathbf{o}_{t}$, and robot state $\mathbf{s}_{t}$. Specifically, different from GR-3 which learns binary discrete gripper actions, each action $a_t$ is a vector of length 88, consisting of: 1) arm joint actions (7 DoFs per arm), 2) arm end-effector poses (6D per arm), 3) hand joint actions (16 active DoFs per hand), and 4) fingertip positions (3D per finger).

\subsection{Training Recipe}
\begin{wrapfigure}[15]{r}{0.4\textwidth}

  \vspace*{-20pt}

  \centering
  \includegraphics[width=0.36\textwidth]{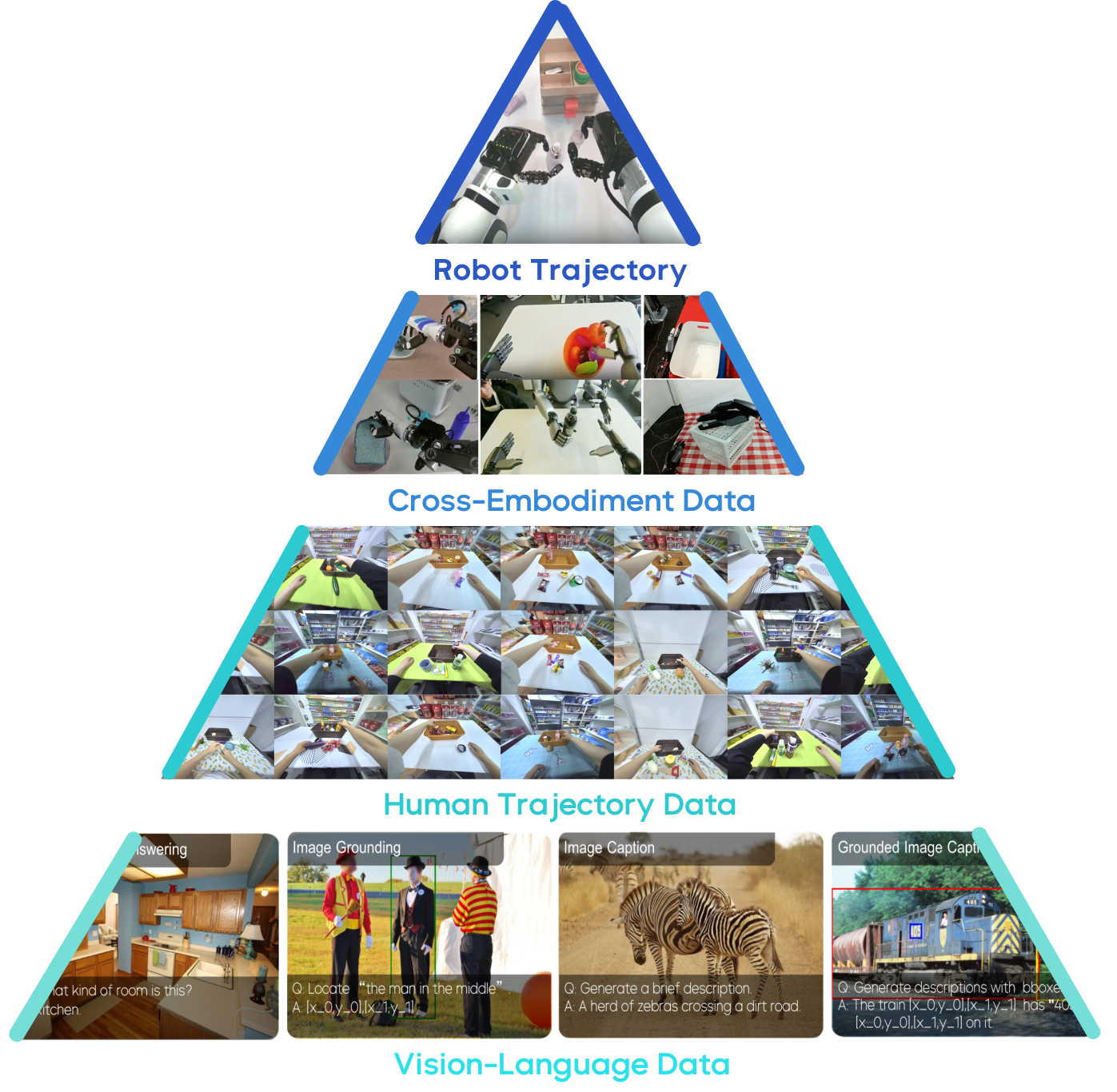}  %
  \caption{Data Pyramid of \name{}. 
  }
  \label{fig:dataset}
\end{wrapfigure}
 We train GR-Dexter using a mixture of three distinct data sources: web-scale vision-language data, cross-embodiment real-robot data, and human trajectory data.

 \paragraph{Vision-language data}
 We reuse the vision-language dataset from GR-3, which covers a wide spectrum of tasks including image captioning, visual question answering, image grounding, and interleaved grounded image captioning. The robot trajectory data are used to train both the VLM backbone and the action DiT using the flow-matching objective. The vision-language data are used to train only the VLM backbone via the next-token-prediction objective. For simplicity, we dynamically mix vision-language data with robot trajectories across mini-batches. As a result, the co-training objective is the sum of the next-token-prediction loss and the flow-matching loss.

 \paragraph{Cross-embodiment data}
 Collecting large-scale teleoperation data on our high-DoF Byte-Dexter platform is constrained by hardware availability and the scarcity of skilled teleoperators. To mitigate this, we leverage existing open-source bi-manual humanoid datasets. Specifically, we select three dual-arm dexterous manipulation datasets that encompass diverse embodiments and task settings: Fourier ActionNet Dataset~\cite{fourier2025actionnet}, 
 which contains around 140 hours of diverse humanoid bimanual manipulation data using Fourier 6-DoF hands; 
 OpenLoong Baihu Dataset~\cite{openloong2025baihu}, which features over 100k robot trajectory data across multiple robot embodiment; 
 RoboMIND~\cite{wu2025robomind}, which includes 107k demonstration trajectories across 479 diverse tasks involving 96 object classes.

\paragraph{Human trajectories}
While cross-embodiment robot data offers accurate robot state information, the scale and diversity of tasks are inevitably limited by hardware costs. Crowdsourcing human demonstrations via easily accessible VR devices offers a promising solution to scale up data quantity and diversity. We adopt the human trajectory data (over 800 hours of egocentric video with paired 3D hand and finger tracking data) and supplement it with additional data collected using Pico VR devices.

To handle the structural differences across datasets, we mask out unavailable or unreliable action dimensions (e.g., specific joints not present in the target embodiment).

\subsection{Cross-Embodiment Motion Retargeting and Transferring}

Transferring dexterous manipulation skills across heterogeneous embodiments and from human demonstrations requires careful calibration of both visual perceptions and action spaces. We address this challenge with a unified preprocessing and retargeting pipeline that aligns visual geometry, kinematics, and trajectory quality across all data sources.

\paragraph{Transferring cross-embodiment trajectories}
We first standardize camera observations across datasets. 
All images are resized and cropped to a standardized format where robot arms, dexterous hands, and object sizes at a similar scale. Such a process can be easily achieved manually for each dataset once and applied to all. 
Trajectories then undergo strict quality control and only high-quality trajectories are maintained. We then perform careful retargeting to ByteDexter V2 hand by aligning the fingertips. 
This fingertip-centric alignment preserves task-relevant contact geometry while remaining agnostic to joint-level discrepancies. The resulting trajectories are then resampled by task category to produce a balanced cross-embodiment training corpus.

\paragraph{Transferring human trajectories}
Human demonstrations pose additional challenges beyond cross-robot transfer. The kinematic gap between human and robotic hands is substantial: VR data collection introduces ego-motion due to head-mounted cameras, and single-frame hand pose estimation commonly leads to temporal jitter and inconsistency—especially during rapid motion or partial occlusion.
We first perform careful filtering based on hand visibility and velocity. 
Next, human trajectories are mapped into the same visual and kinematic representation as robot data similar to the cross-embodiment data cleaning process,
enabling seamless integration into the GR-Dexter training pipeline.

%% file: sections/experiments.tex
\section{Experiments}
\label{sec:experiments}

We conduct extensive real-world experiments to evaluate the performance of \name{} on long-horizon bimanual manipulation and generalizable pick-and-place tasks. We evaluate \name{} in: (1) challenging dexterous tool use tasks, (2) long-horizon task execution, and (3) OOD scenarios with novel relative spatial configurations, unseen objects, and unseen instructions.

As a summary of our results, we show that \name{} achieves strong long-horizon manipulation capabilities with 21-DoF ByteDexter V2 hands, and generalizes better to unseen settings given our curated data pyramid (robot trajectories, vision–language data, cross-embodiment demonstrations, and human trajectories).

\subsection{Long-Horizon Dexterous Manipulation}

\begin{figure}[t]
  \centering
  \includegraphics[width=1\linewidth]{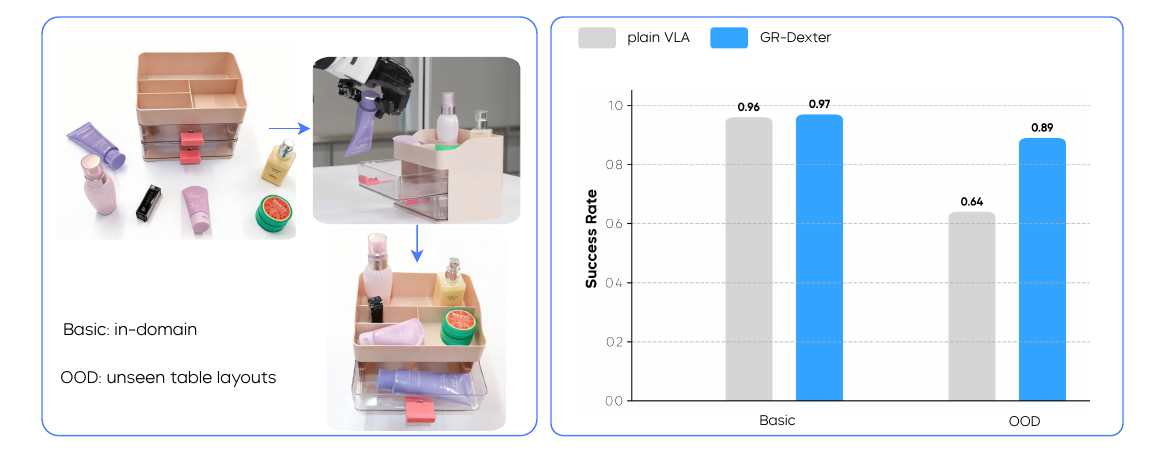}
  \caption{\textbf{Experiment Settings and Results} of Makeup Decluttering.}
  \label{fig:makeup_result}
\end{figure}

We first test the long-horizon manipulation capabilities of \name{} on a makeup decluttering task involving long-sequence manipulation of items with diverse shapes and sizes, as well as articulated objects like drawers. The task requires coordinated bimanual manipulation and fine-grained skills. 
We collected approximately 20 hours of teleoperated robot trajectories.
We train \name{} with a co-training strategy with both vision-language data and teleoperation robot trajectories.
We also compare \name{} against a plain VLA baseline trained with only robot data.
During policy rollout, the robot is sequentially prompted with natural-language subtask descriptions (six items; one instruction per item) until the task is completed. Each subtask execution starts from the robot's home pose. We report task performance using the average \textbf{success rate} across multiple evaluation trials. Fig.~\ref{fig:makeup_result} summarizes success rates for plain VLA and \name{}.

\paragraph{Basic Settings}
In the basic settings, the relative spatial configurations (layouts) of objects are present in the training data. Here, plain VLA has a comparable performance with \name{}, achieving 0.96 and 0.97 success rates correspondingly. This shows that co-training preserves the strong in-domain capability of the teleop-only baseline

\paragraph{Our-of-Distribution Settings}
Here, the relative spatial configurations of objects are novel at test time. We evaluate on five unseen layouts while keeping the instruction order the same as Basic. In OOD settings, the performance of the plain VLA drops to 0.64, whereas \name{} improves substantially to 0.89. These results indicate that co-training with vision-language data significantly enhances generalization to unseen spatial layouts, while maintaining in-domain performance.

\paragraph{Additional Qualitative Results}
Besides the makeup decluttering task, we further show \name{} is capable of more complex long-horizon manipulation with tool use. Specifically, we consider two tasks:
\begin{itemize}[leftmargin=*]
  \item \textbf{Vacuuming:} the robot learns a stable four-finger grasp to hold the tabletop vacuum while using the thumb to press the power button (on/off). Next, it presses again to increase power, then sweeps to clear confetti.
  \item \textbf{Bread serving:} the robot learns to stably grasp food tongs to retrieve a croissant from a pastry container while the other hand holds a plate. It then releases the tongs and places the croissant onto the plate with precise, compliant manipulation.
\end{itemize}
We observe \name{} performs both tasks reliably across time. Please visit the \href{https://byte-dexter.github.io/gr-dexter/}{project page} for detailed videos.

\subsection{Generalizable Pick-and-Place}

\begin{figure}[h]
    \centering
        \centering
        \includegraphics[width=0.95\linewidth]{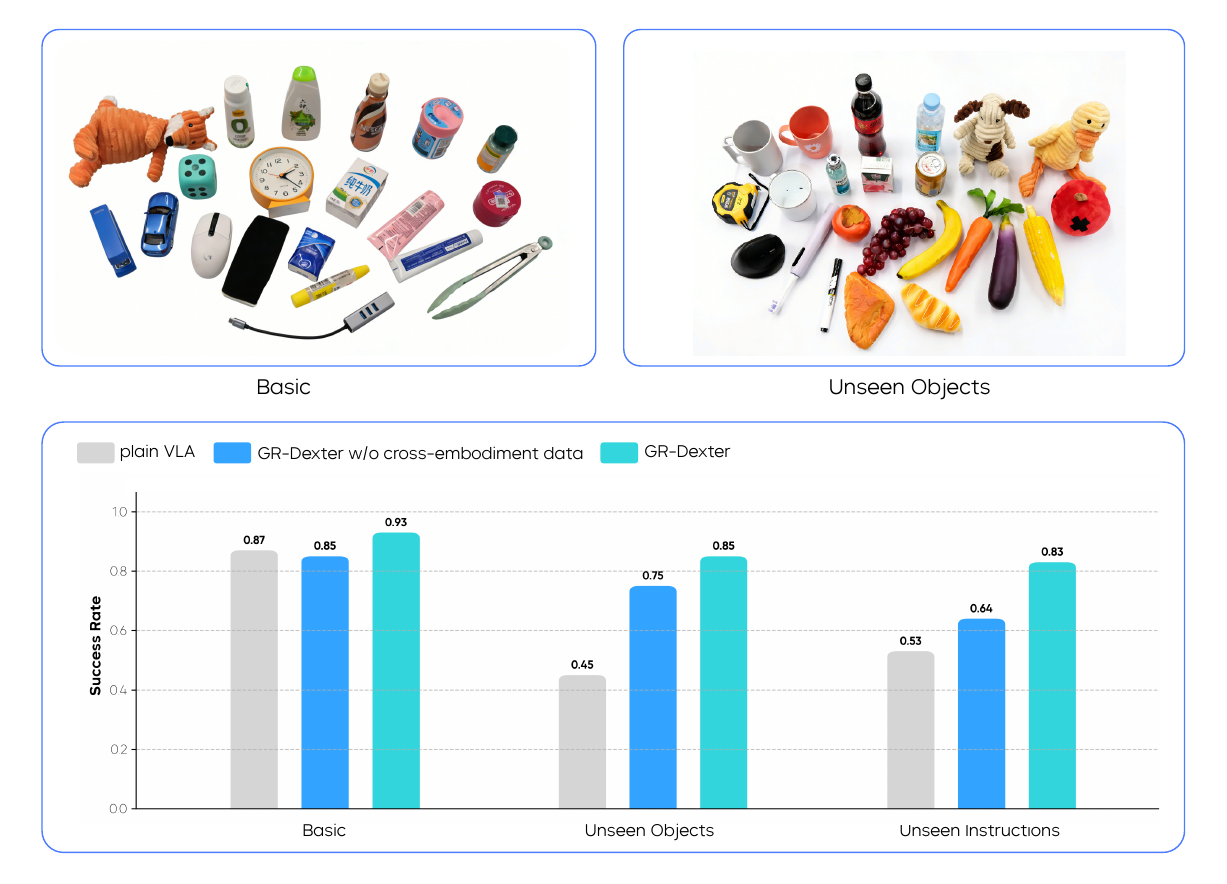}
        \caption{\textbf{Experiment settings and Results} of Generalizable Pick-and-Place.}
        \label{fig:ppa_result}
\end{figure}

We evaluate the generalization capabilities of \name{} on a pick-and-place task. We collected approximately 20 hours of robot trajectories with 20 objects for training (Fig.~\ref{fig:ppa_result}). We compare three models: plain VLA, \name{} without cross-embodiment data, and \name{}. We evaluate performance using task success rate. During policy rollout, the model is prompted with a natural-language instruction specifying a target object. A trial is considered successful if the robot picks up the target object and places it into the container. For each evaluation batch, we keep the object layout fixed across rollouts for all policies.

\paragraph{Basic Settings}
We construct 10 evaluation batches using seen objects, with five objects per batch.
We observe that in the in-domain Basic setting, plain VLA reaches 0.87, \name{} (w/o cross-embodiment data) reaches 0.85, and GR-Dexter achieves the best performance at 0.93. We find the results interesting because: (1) \name{} w/o cross-embodiment data performs slightly worse than plain VLA, as in the in-distribution setting, VL data gives no additional information but makes optimization more challenging; (2) with \textit{cross-embodiment} data, \name{} significantly outperforms the two baselines, which suggests after careful data processing and alignment, larger scale cross-embodiment training for the action expert can improve the overall robustness and performance of \name{}.

\paragraph{Unseen Objects and Instructions}
For unseen objects, we select 23 unseen objects and construct 10 evaluation batches, with five objects per batch. Besides, we construct 5 evaluation batches using seen and unseen objects, and prompt the model with unseen language instructions. In both settings, we observe that (1) the performance of plain VLA drops significantly; (2) VLM co-training largely improves the robustness and generalization of \name{}, but empirically, \name{} w/o cross-embodiment data still suffer from inaccurate grasping; (3) with carefully filtered and aligned cross-embodiment co-training, \name{} demonstrates strong generalization capabilities to both unseen objects and instructions, achieving a final success rate of 0.85 and 0.83 respectively.
These gains are consistent with the qualitative examples in Fig.~\ref{fig:cross_embodied}, where GR-Dexter successfully grasps unseen objects by leveraging skills learned from cross-embodiment data, and Fig.~\ref{fig:unseen_ins}, where it correctly interprets and executes previously unseen instructions.

%% file: sections/related_works.tex
\section{Related Works}
\label{sec:related_works}

\subsection{Dexterous Robotic Hands}

Recent years have witnessed rapid progress in multi-fingered dexterous robotic hands, particularly in industrial robotics and manufacturing~\cite{simlab2015allegro,shaw2023leap,wuthrich2020trifinger,liu2008multisensory,deimel2016rbo}. More than ten industrial and startup players have commercialized such hands, including Unitree~\cite{dex5}, AgiBot~\cite{omnihandpro}, and Fourier~\cite{fdh6}. Most commercial designs use a relatively small number of active DoFs (typically 6), while a smaller subset targets 12; only a few exceed 12 active DoFs. The SharpaWave hand is among the most highly actuated commercial systems, featuring motor direct-drive actuation with 22 independently actuated DoFs, and is one of the most integrated dexterous robotic hands to date~\cite{sharpa2025}. A representative tendon-driven platform is the Shadow Hand~\cite{shadow}. More recently, Dexcel Robotics released the Apex Hand~\cite{apex2025}, which provides 21 DoFs (16 independently actuated) and dense tactile sensing over the fingertips, phalanges, and palm. A third transmission paradigm is linkage-driven actuation, exemplified by the ILDA hand~\cite{KimILDA2021} and the ByteDexter V1 hand~\cite{wen2025dexterous}; both provide 20 DoFs, 15 of which are independently actuated. Compared with tendon-driven designs, linkage-driven hands can improve durability and force transparency, simplify maintenance, and enable compact actuator integration within the palm---allowing the hand to function as a self-contained, modular unit without external actuation components.

Building upon the ByteDexter V1 design, this work presents an upgraded version that increases the total DoFs by one while achieving a more compact form factor. The new design further incorporates high-density piezoresistive tactile sensors covering the fingertips, enhancing its suitability for dexterous manipulation and fine contact-rich tasks.

\subsection{VLA Models for Dexterous Hand Manipulation}

VLA models have emerged as foundation policies for generalist manipulation, demonstrating strong instruction-following and long-horizon capabilities~\cite{gr3_2025,pi05,grrl_2025,brohan2023rt,o2024open,team2024octo, bousmalis2023robocat,huang2023voxposer,jiang2023vima,liu2023roboflamingo,ma2024robomamba,belkhale2024hip}. Although these models are advancing rapidly, their application to dexterous, multi-fingered hands is limited. Integrating anthropomorphic hands into bimanual manipulation substantially increases control dimensionality—often by several dozen DoFs relative to gripper-based setups—thereby raising the demands on both modeling and data. VLA performance depends critically on the diversity and quality of demonstration trajectories; however, large-scale teleoperated dexterous-hand datasets remain scarce. 

Recent work suggests that pretraining on human videos can partially mitigate this bottleneck by transferring dexterous manipulation priors to robot policies~\cite{beingbeyond2025beingh0,shaw2023videodex,wang2023mimicplay,radosavovic2023real,li2025vitra,hoque2025egodex,qin2022dexmv,sivakumar2022robotic,arunachalam2023dexterous,ma2023vip,ma2023liv,bahl2022human,he2024dest}. GR00T N1 follows this paradigm for humanoid robots. It combines pre-training and post-training on heterogeneous data sources, including real-robot, synthetic, and human video datasets, and has demonstrated effectiveness on Fourier GR-1 humanoids equipped with 6-DoF dexterous hands~\cite{gr00tn1_2025}. Hierarchical approaches also decouple planning from control by using a pretrained vision–language model for task planning~\cite{li2023vision, li2024cogact, wen2025dexvla}. Low-level execution is implemented either with a diffusion transformer (DiT) that outputs action chunks conditioned on grasp instructions and bounding boxes~\cite{zhong2025dexgraspvla,li2025reinforcement,bharadhwaj2024roboagent,chi2024diffusionpolicy,xian2023chaineddiffuser} or with RL-based controllers that track the generated trajectories~\cite{debakker2025scaffold,luo2024precise,luo2025tsl,lu2025vla}. 
Going beyond prior works, \name{} extends VLA models to 21-DoF dexterous hand manipulation by co-training on a mixture of teleoperated robot trajectories, vision-language data, cross-embodiment data, and human trajectories, enabling long-horizon dexterous performance.

\subsection{Bimanual Dexterous Manipulation Dataset}

Most existing dexterous manipulation datasets focus on single-hand grasping~\cite{chao2021dexycb,feix2016,brohan2022rt,zeng2021transporter}. They often emphasize static grasps on isolated objects and typically lack language supervision and whole-body arm–hand trajectories, making them less suitable for bimanual manipulation that requires coordinated dual-arm control and long-horizon task execution. With the recent rise of VLA models as foundation policies, open-source datasets have grown rapidly, broadly following two paradigms:

\paragraph{Teleoperated Robot Data} Operators control humanoid robots through teleoperation interfaces (e.g., VR controllers or data gloves) to perform predefined tasks (e.g., RoboMIND~\cite{wu2025robomind}, OpenLoong Baihu~\cite{openloong2025baihu}). These datasets provide high-fidelity joint states and actions, synchronized visual observations, and often high-quality language annotations. However, task and environmental diversity is frequently limited by hardware cost and operational complexity. Moreover, unlike grippers, dexterous hands vary substantially across platforms; the resulting kinematic discrepancies further complicate cross-embodiment transfer.

\paragraph{Human Trajectory Data} Egocentric recordings collected with VR devices and wearable cameras~\cite{grauman2022ego4d,banerjee2024hot3d,cui2025dexgraspvla,shao2024teleop,mandlekar2018roboturk} scale well and cover diverse scenes and tasks. However, the large embodiment gap between human and robot hands, together with noisy state estimation, complicates both learning and retargeting for robot control.

In this technical report, we address these challenges by constructing a unified dataset for co-training. We combine curated subsets from public bimanual manipulation datasets with proprietary robot teleoperation trajectories and human demonstrations. Using a standardized pipeline for data cleaning, retargeting, and post-processing, we produce a dataset that balances the precision of robot trajectories with the semantic and environmental diversity of human demonstrations.

%% file: sections/limitation_future_work.tex
\newpage
\section{Limitations \& Conclusions}
\label{sect:limitation_future_work}

\paragraph{Limitations and future work}
Our current system has several limitations that suggest clear directions for future work: (1) on the human side, we leverage only a few hundred hours of human trajectories, leaving substantial complementary egocentric human data untapped; and (2) the robot’s hand and arm are controlled separately, which can hinder tight hand–arm coordination in contact-rich dexterous behaviors. Going forward, it is crucial to further improve the pre-training scale by exploiting diverse and more accessible cross-embodiment trajectories, and building embodiment-agnostic control abstractions.

\paragraph{Conclusions}
We introduce \name{}, a hardware-model-data approach that extends VLA-based generalist manipulation to high-DoF bimanual dexterous-hand robots. On the hardware side, we present ByteDexter V2, a compact anthropomorphic hand designed for dexterous manipulation. On the data side, we develop an intuitive bimanual teleoperation pipeline that makes collecting high-quality demonstrations feasible at this dimensionality. Building on these components, \name{} co-trains a VLA policy for a 56-DoF bimanual robot using teleoperated robot trajectories together with vision-language data and a carefully curated dataset of cross-embodiment demonstrations and human trajectories. In real-world evaluations on long-horizon everyday manipulation and generalizable pick-and-place, \name{} achieves strong in-domain performance and improved robustness to unseen objects and instructions. These results suggest that combining practical dexterous hardware with scalable data collection and cross-embodiment supervision is a promising path toward generalist dexterous-hand manipulation.

%% file: sections/contribution.tex
\clearpage
\section*{Contributions and Acknowledgements}
\phantomsection
\addcontentsline{toc}{section}{Contributions and Acknowledgments}
\label{sec:contributions}

\paragraph{Hardware} 
Jiajun Zhang, Zhigang Han, Guangzeng Chen, Zhongren Cui, Min Du, Hao Niu, Yang Gou, Zeyu Ren, Wenlei Liu, Mingyu Lei, Liwei Zheng, Xiao Zhang

\paragraph{Control and System} 
Liqun Huang, Ruoshi Wen, Zhongren Cui, Zhengming Zhu, Zeyu Ren, Zhuohang Li, Haoxiang Zhang

\paragraph{Model and Data}
Haixin Shi, Wei Xu, Ruoshi Wen, Liqun Huang, Weiheng Zhong, Yutao Ouyang, Zhuohang Li, Yunfei Li, Yifei Zhou, Yuxiao Liu, Xiao Ma

\paragraph{Supervisor}     
Hang Li

We thank Jinming Guo, Zetian Li, and Degong Yang for their help on data curation and system maintenance. We are sincerely grateful to all teleoperators and annotators for their dedicated efforts on data collection and annotation.

 This work is presented for research purposes only. The technology described in this paper will not be incorporated into any ByteDance product.

%% file: sections/appendix.tex
\section*{Grasping Capability}
\label{sec:appendix}

\begin{figure}[h]
    \centering
    \includegraphics[width=\linewidth]{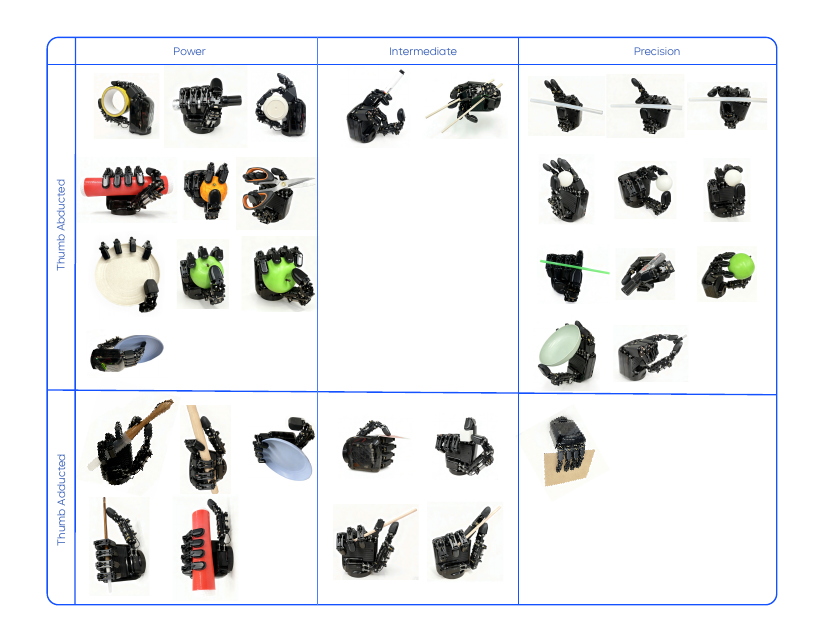} 
    \caption{\textbf{ByteDexter V2} demonstrates human‑like grasping, with a workspace that supports 33 grasp types.}
    \label{fig:hand_feix} 
\end{figure}